# Dynamic EEG-fMRI mapping: Revealing the relationship between brain connectivity and cognitive state

Guiran Liu*
San Francisco State University, College of Science & Engineering (CoSE)
San Francisco, United States
gliu@sfsu.edu

Binrong Zhu
San Francisco State University, College of Science & Engineering (CoSE)
San Francisco, United States
bzhu2@sfsu.edu

*Abstract*-: **This study investigated the dynamic connectivity patterns between EEG and fMRI modalities, contributing to our understanding of brain network interactions. By employing a comprehensive approach that integrated static and dynamic analyses of EEG-fMRI data, we were able to uncover distinct connectivity states and characterize their temporal fluctuations. The results revealed modular organization within the intrinsic connectivity networks (ICNs) of the brain, highlighting the significant roles of sensory systems and the default mode network. The use of a sliding window technique allowed us to assess how functional connectivity varies over time, further elucidating the transient nature of brain connectivity. Additionally, our findings align with previous literature, reinforcing the notion that cognitive states can be effectively identified through short-duration data, specifically within the 30-60 second timeframe. The established relationships between connectivity strength and cognitive processes, particularly during different visual states, underscore the relevance of our approach for future research into brain dynamics. Overall, this study not only enhances our understanding of the interplay between EEG and fMRI signals but also paves the way for further exploration into the neural correlates of cognitive functions and their implications in clinical settings. Future research should focus on refining these methodologies and exploring their applications in various cognitive and clinical contexts.**

*Keywords*- **EEG-fMRI Integration; Dynamic Connectivity ; Functional Brain Networks；Cognitive States；**

## I. Introduction

Graph theory-based analysis serves as a robust tool to quantitatively describe the topological characteristics of brain networks[1]. This method has been widely applied in brain imaging research, revealing an "economical" small-world organization in brain networks, where an efficient balance between network cost and efficiency is maintained. Additionally, the brain connectome exhibits a modular and rich-club organization, which is crucial for the transmission and communication of neural signals. Studies have shown that brain disorders can alter graph metrics and the structure of brain networks. However, most research has focused on single-modality brain imaging data, limiting a more comprehensive understanding of brain connectivity[2]. Thus, integrating multiple modalities, especially EEG and fMRI, offers a deeper understanding of brain dynamics and network topology[3].

EEG and fMRI are two complementary imaging techniques used to study brain activity[4]. fMRI has good spatial resolution and captures hemodynamic responses across the entire brain, but its temporal resolution is relatively low. On the other hand, EEG measures cortical electrical activity with high temporal resolution, but it has lower spatial resolution. Combining both signals is an effective way to study brain dynamics across broader spatial and temporal scales. Previous research has found that the low-frequency connectivity detected by EEG closely resembles brain connectivity observed in fMRI, shedding light on the electrophysiological basis of functional brain connectivity[5].

Neurofeedback (NF) is a technique that helps individuals regulate brain activity through real-time feedback and is commonly used in rehabilitation and the treatment of mental health disorders[6]. Both EEG and fMRI are non-invasive functional brain imaging methods used in neurofeedback. EEG offers millisecond-level temporal resolution, enabling real-time monitoring of electrical brain activity, while fMRI reflects neurovascular activity through BOLD signals and provides high spatial resolution. Although fMRI has been widely used in neurofeedback, its high cost and operational complexity limit frequent use in clinical settings. As a result, EEG-based neurofeedback has gained attention due to its convenience and flexibility. In recent years, simultaneous recording of EEG and fMRI has been used to explore connections in different brain states, further expanding the applications of multimodal measurements[7].

In this study, the graph theory properties of multimodal EEG-fMRI brain connectivity were explored. By using simultaneously collected open- and closed-eye data, brain graphs were constructed in both static and dynamic states, combining the high spatial resolution of fMRI and the high temporal resolution of EEG. This approach offers a new framework that integrates the different information from EEG and fMRI within a graph theory model, providing a more comprehensive representation of the brain's dynamic connectivity organization[8].

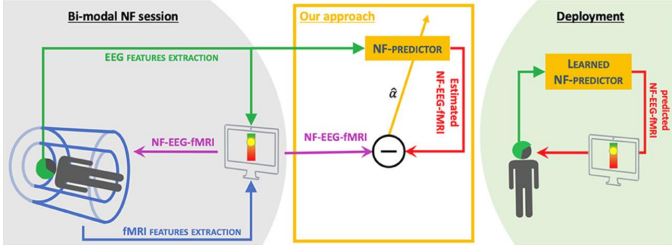

Figure 1. A Method for Learning NF Predictors from Bimodal Neurofeedback Sessions

## II. Method

### A. Participator

We recruited 25 healthy participants (age: 29 ± 8 years; 8 females) through advertisements and word of mouth. All individuals had normal or corrected-to-normal vision and hearing.[9] Prior to inclusion in the study, participants underwent screening to ensure the absence of any DSM-IV Axis I or Axis II psychopathology, assessed through a structured interview, as well as no history of neurological disorders. Each participant provided informed written consent for the study, and they were compensated for their participation. The experimental design and details regarding simultaneous data collection have been previously described.

### B. Experimental design

EEG-fMRI data were recorded simultaneously while participants first rested with their eyes closed (for 8.5 minutes), followed by resting with their eyes open (for another 8.5 minutes). During each session, participants were instructed to relax, remain still and quiet, and stay awake throughout the recording[10].

### C. Eeg acquisition

EEG data were collected using a 32-channel system compatible with MR environments and a specialized electrode cap. The Ag/AgCl electrodes were arranged based on the international 10-20 standard. Separate channels recorded ECG and EOG signals, limiting the number of scalp electrodes to 30. The reference electrode was positioned at FCz. Conductive paste ensured that the impedance of each electrode remained below 5 kΩ. EEG signals were sampled at a rate of 5 kHz, and synchronization between the EEG amplifier and the fMRI system was achieved using an internal device to avoid timing discrepancies.

### D. Fmri acquisition

Functional MRI brain images were obtained using a 1.5 T Siemens Sonata scanner with a T2*-weighted echo-planar imaging sequence. The key parameters included: repetition time (TR) of 2 seconds, echo time (TE) of 39 milliseconds, field of view set at 224 mm, acquisition matrix of 64 × 64, flip angle of 80 degrees, voxel dimensions of 3.5 × 3.5 × 3 mm, and a 1 mm gap between slices. A total of 27 slices were acquired in ascending order, with 256 volumes recorded for both the eyes-open and eyes-closed sessions.

### E. Functional magnetic resonance imaging

The fMRI data underwent preprocessing using SPM5 software. Images were realigned and spatially normalized to MNI space, with a voxel size resampled to 3 × 3 × 3 mm and smoothed using a Gaussian kernel (FWHM of 5 × 5 × 5 mm).

A spatial group independent component analysis (ICA) was performed on the fMRI data from all participants for both conditions (eyes-open and eyes-closed). The data for each participant were reduced using principal component analysis (PCA), retaining 120 principal components. Subsequently, the simplified data were decomposed into 100 independent components using the Infomax algorithm. The stability of these components was confirmed through ICASSO analysis, which involved 10 iterations. The independent components (IC) and their corresponding time courses (TC) were reconstructed. Fifty-four ICs were identified as intrinsic connectivity networks (ICN), excluding those associated with physiological noise, motion-related artifacts, or imaging artifacts. The evaluation of these components was based on the expectation that ICNs would exhibit peak activation in gray matter, with minimal spatial overlap with known vascular, ventricular, motion, and susceptibility artifacts, and primarily low-frequency fluctuations (less than 0.1 Hz). Following established protocols, the TCs for the 54 ICs underwent additional post-processing, which included detrending for linear, quadratic, and cubic trends, multiple regression to adjust for six parameters and their temporal derivatives, removal of detected outliers, and bandpass filtering within the range of [0.01–0.10 Hz]. Finally, the ICA time course matrices were extracted for each participant under both conditions (eyes-open and eyes-closed), resulting in a time matrix of dimensions [time: 256 × ICN: 54]. All participants in the study were healthy volunteers, right-handed, and had no prior experience with neurofeedback experiments. They provided written informed consent before taking part in the study. Following a specialized calibration session, each participant underwent three neurofeedback motor imagery sessions, each lasting 320 seconds. Each session consisted of eight blocks that alternated between 20 seconds of rest, with eyes open, and 20 seconds of motor imagery for the right hand. Neurofeedback data from thirteen participants were displayed as one-dimensional (1D) in Figure 2 (left), while data from twelve participants were shown as two-dimensional (2D) in Figure 2 (middle). In both scenarios, the goal was to guide a ball into a deep blue area.

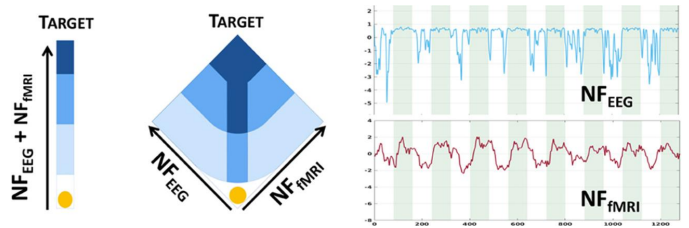

Figure 2. The dual-modal neurofeedback metaphor presented during the conference

## F. EEG-fMRI brain map was constructed

The correlation matrix R is formed by utilizing elements that represent Pearson correlation coefficients ($r_{ij}$). These coefficients are derived from the spectral time series of 30 EEG electrodes and the time series from 54 fMRI independent components (ICs). This procedure is conducted for five EEG frequency bands across two conditions (Eyes Open (EO) and Eyes Closed (EC)). To analyze the relationship between EEG and fMRI signals, the EEG power time series undergoes convolution with a standard hemodynamic response function (HRF) to account for the latency in hemodynamic responses, as supported by earlier studies.

The undirected static connectivity graph for EEG-fMRI is constructed from each N×N correlation matrix R (where N=84 in this study, comprising 30 EEG electrodes and 54 fMRI brain components). This graph includes negative correlations, as well as weighted positive (W+) and negative (W−) connections. In the positive connection graph, negative correlation values in RRR are replaced by zero, while the positive correlation values remain unchanged. Conversely, in the negative connection graph, positive correlations are set to zero, and the absolute values of negative correlations in R are maintained.

$$w_{ij}^+ = \begin{cases} r_{ij} & if\, r_{ij} > 0 \\ 0 & if\, r_{ij} \leq 0 \end{cases} \quad (1)$$

$$w_{ij}^- = \begin{cases} |r_{ij}| & if\, r_{ij} < 0 \\ 0 & if\, r_{ij} \geq 0 \end{cases} \quad (2)$$

Dynamic EEG-fMRI graph analysis involves the calculation of correlation matrices (256 × 84; width L=20 TRs, with a step size of 1 TR) using a continuous sliding window across the matrix EFEFEF. The first 30 columns represent the EEG electrodes, while the next 54 columns correspond to the fMRI independent components (ICs). This method results in 237 EEG-fMRI correlation matrices (237 = 256 − 20 + 1) calculated from 237 windows. As in the static analysis, both positive and negative connection graphs are examined separately. The framework for constructing static and dynamic concurrent EEG-fMRI multimodal brain maps is depicted in Figure 3.

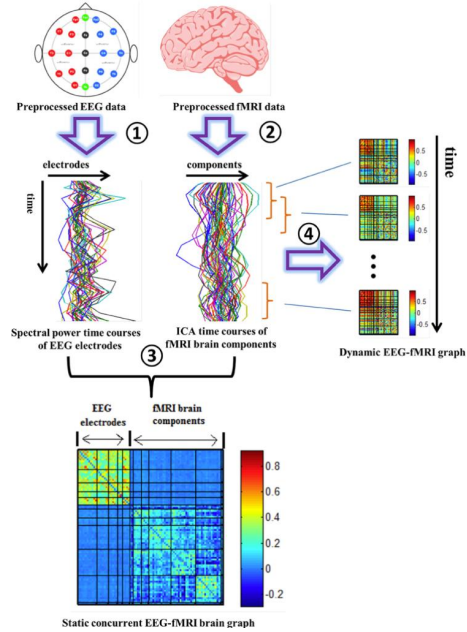

Figure 3. The process of constructing concurrent EEG-fMRI multimodal brain maps involves several key steps.

We utilized a window width of 20 TRs (40 seconds) based on research indicating that cognitive states can be accurately identified with data from short periods of 30 to 60 seconds. Studies have shown that non-stationary fluctuations in functional connectivity can be detected with a 40-second window. Variations in brain connectivity are not particularly sensitive to specific window lengths in the range of 10-20 TRs (20-40 seconds). Our previous work has demonstrated that shorter time windows reduce the number of statistically significant correlations in brain connectivity and increase variability. A sliding window size of approximately 22 TRs (44 seconds) strikes a good balance between resolving dynamics and the quality of connectivity estimates.

We employed three key metrics—connection strength (CS), clustering coefficient (CC), and global efficiency (GE)—to assess brain network functionality. The Brain Connectivity Toolbox was used to extract global and node-level values for both static and dynamic graphs. We computed the variance of dynamic metrics across 237 time windows and low-frequency (0-0.025 Hz) fluctuation amplitudes. For statistical analysis, a 5 (frequency bands) × 2 (eye conditions) repeated measures ANOVA and paired t-tests were conducted on the static and dynamic measurements.

## G. Calculate the equation of the graph measurement

Let G denote the collection of all nodes in the weighted graph W, where NNN (in this study, N=84) signifies the total number of nodes. The connectivity strength for node i is defined as follows:

$$CS_i = \sum_{j \in G} w_{ij} \quad (3)$$

The connectivity strength at the global level of the graph is the average connectivity strength of all nodes within the graph:

$$CS_{net} = \frac{1}{N} \sum_{i \in G} CS_i \quad (4)$$

The clustering coefficient at the node level is calculated using the following formula:

$$CC_i = \frac{1}{CS_i(CS_i - 1)} \sum_{j,k \in G} (w_{ij} w_{ik} w_{jk})^{1/3} \quad (5)$$

The clustering coefficient for the entire graph is calculated as the average of the clustering coefficients across all nodes within the graph:

$$CC_{net} = \frac{1}{N} \sum_{i \in G} CC_i \quad (6)$$

The global efficiency of node i is defined as:

$$GE_i = \frac{\sum_{j \in N, j \neq i} (d_{ij})^{-1}}{N - 1} \quad (7)$$

Where:

$$d_{ij} = \sum_{a_w \in g_w^{i \leftrightarrow j}} f(w_{uv}) \quad (8)$$

In this context, f refers to a mapping (specifically the inverse) that relates weight to length, while $g_{i \leftrightarrow j}$ represents the shortest weighted path connecting nodes iii and j. The global efficiency of the graph is determined by averaging the global efficiencies of all nodes in the graph.

### H. Detect connection status

Recent research using functional magnetic resonance imaging has shown that fluctuations in time-varying functional brain connectivity lead to distinct, well-organized patterns, known as connectivity states, which can appear or disappear over time. In this study, we apply a previously developed method to identify the connectivity states of dynamic EEG-fMRI graphs for each individual.

Initially, we compute the node-level connectivity strength for each time-varying EEG-fMRI graph. To evaluate how EEG-fMRI network patterns are related across different time windows, we create a new correlation matrix based on the connectivity strengths between nodes across all time windows. This involves examining each pair of time windows across the 84 nodes.

Modular community structure is a prevalent characteristic of complex networks. Modularity assesses the quality of grouping nodes into communities, where the modules in the correlation matrix correspond to groups of time windows displaying similar brain connectivity patterns. We analyze the modular organization of this correlation matrix using a modularity algorithm. The number of modules reflects the different connectivity states present within the dynamic EEG-fMRI graphs. Lastly, we average the EEG-fMRI brain graphs belonging to the same module to derive the graph representing that connectivity state.

### III. Result
#### A. Spatial maps of fMRI brain components

Figure 4A illustrates the spatial maps for the 54 independent component networks (ICNs) identified through independent component analysis (ICA). These ICNs are classified into categories such as subcortical (SC), auditory (AUD), sensorimotor (SM), visual (VIS), cognitive control, default mode (DM), and cerebellar (CB) components, based on their anatomical and presumed functional attributes. The identified ICNs share similarities with those found in earlier high model-order ICA decompositions, with a subset being associated with metacognitive functions as revealed in relevant studies.

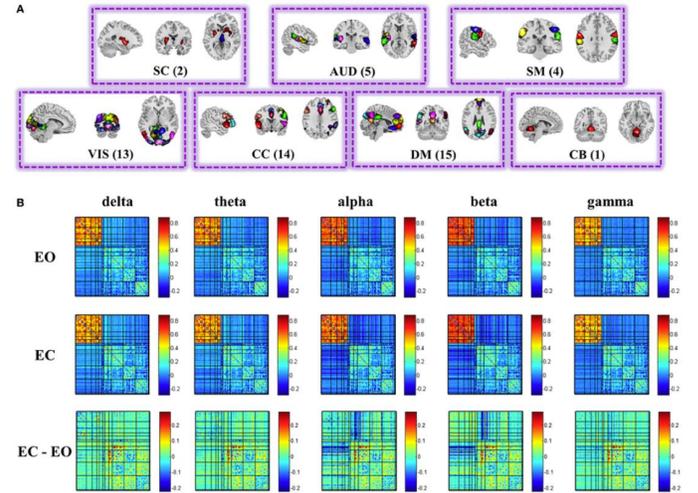

Figure 4 illustrates the spatial maps of 54 independent components (A) and the group average static EEG-fMRI brain maps highlighting structural differences across five frequency bands in open-eye, closed-eye, and eye conditions (B).

Figure 4B illustrates the static connectivity structure between graph nodes (ICNs and EEG channels), calculated and averaged over fMRI time series and EEG spectral power across five frequency bands (delta, theta, alpha, beta, low gamma) for more than 25 participants in both open-eye and closed-eye conditions. The connectivity patterns within fMRI ICNs show modular organization in sensory systems and default mode regions, along with anti-correlations between these areas.

### IV. Conclusion

This study investigated the dynamic connectivity patterns between EEG and fMRI modalities, contributing to our understanding of brain network interactions. By employing a comprehensive approach that integrated static and dynamic analyses of EEG-fMRI data, we were able to uncover distinct connectivity states and characterize their temporal fluctuations.

The results revealed modular organization within the intrinsic connectivity networks (ICNs) of the brain, highlighting the significant roles of sensory systems and the default mode network. The use of a sliding window technique allowed us to assess how functional connectivity varies over time, further elucidating the transient nature of brain connectivity.

Additionally, our findings align with previous literature, reinforcing the notion that cognitive states can be effectively identified through short-duration data, specifically within the 30-60 second timeframe. The established relationships between connectivity strength and cognitive processes, particularly during different visual states, underscore the relevance of our approach for future research into brain dynamics.

Overall, this study not only enhances our understanding of the interplay between EEG and fMRI signals but also paves the way for further exploration into the neural correlates of cognitive functions and their implications in clinical settings. Future research should focus on refining these methodologies and exploring their applications in various cognitive and clinical contexts.